%% file: main.tex
\documentclass[10pt,twocolumn,letterpaper]{article}

\usepackage{cvpr}
\usepackage{balance}  
\usepackage{url} 
\usepackage{amssymb}
\usepackage{amsmath}
\usepackage{caption}
\usepackage{listings}
\usepackage{paralist} 
\usepackage{booktabs} 
\usepackage{graphicx}
\usepackage{multirow}
\usepackage{array}
\usepackage{microtype} 
\usepackage[shortlabels]{enumitem}
\setlist{topsep=3pt, itemsep=1pt, parsep=0pt, leftmargin=1.4em}

\usepackage{xcolor}
\definecolor{codeframe}{gray}{0.75}
\definecolor{codebg}{gray}{0.97}
\definecolor{codenum}{gray}{0.45}

\lstdefinestyle{python-code}{
  basicstyle=\fontsize{8.5}{10}\ttfamily,
  language=python,
  numbers=left, numberstyle=\tiny\color{codenum}, stepnumber=1,
  xleftmargin=2.2em, framexleftmargin=2.2em,
  aboveskip=8pt, belowskip=2pt,
  breaklines=false,
  frame=single, framerule=0.3pt, rulecolor=\color{codeframe},
  backgroundcolor=\color{codebg},
  columns=fullflexible, keepspaces=true,
}

\lstdefinestyle{mlir-code}{
  basicstyle=\fontsize{8.5}{10}\ttfamily,
  numbers=left, numberstyle=\tiny\color{codenum}, stepnumber=1,
  xleftmargin=2.2em, framexleftmargin=2.2em,
  aboveskip=8pt, belowskip=2pt,
  breaklines=false,
  frame=single, framerule=0.3pt, rulecolor=\color{codeframe},
  backgroundcolor=\color{codebg},
  columns=fullflexible, keepspaces=true,
}

\lstdefinestyle{plain-code}{
  basicstyle=\fontsize{8.5}{10}\ttfamily,
  numbers=none,
  aboveskip=4pt, belowskip=4pt,
  breaklines=false,
  frame=single, framerule=0.3pt, rulecolor=\color{codeframe},
  backgroundcolor=\color{codebg},
  columns=fullflexible, keepspaces=true,
}

\lstdefinestyle{sig-code}{
  basicstyle=\fontsize{8.5}{10}\ttfamily,
  numbers=none,
  aboveskip=6pt, belowskip=6pt,
  breaklines=false,
  frame=single, framerule=0.3pt, rulecolor=\color{codeframe},
  backgroundcolor=\color{codebg},
  columns=fullflexible, keepspaces=true,
}

\newcommand{\op}[1]{{\sf #1}}
\newcommand{\diop}[2]{{\sf #1}.{\sf #2}}  

\cvprfinalcopy

\begin{document}

\title{An MLIR-Based Compilation Method for Large Language Models}

\author{Pengchao Hu\quad Zhibin Xin\quad Yifan Chen\quad Yangyang Zhou\quad Liang Wang\quad Xin Zhang\\
{\tt\footnotesize \{pengchao.hu,zhibin.xin,yifan.chen,yangyang.zhou,liang.wang01,xin.zhang03\}@sophgo.com}\\
Sophgo Inc.
}

\maketitle

\begin{abstract}
  \input{abstract}
\end{abstract}

\section{Introduction}\label{sec:intro}
\input{introduction}

\section{Background}\label{sec:background}
\input{background}

\section{Building Large Models with TopOp}\label{sec:topop}
\input{compiler}

\section{Three-Stage Splitting: Prefill, Prefill\_kv, and Decode}\label{sec:stages}
\input{stages}

\section{Experiments}\label{sec:experiments}
\input{experiments}

\section{Discussion}\label{sec:discussion}
\input{discussion}

\section{Conclusion}\label{sec:conclusion}
\input{conclusion}

{\balance\footnotesize
\bibliographystyle{ieee}
\bibliography{bibliography}
}

\end{document}

%% file: abstract.tex
Large Language Models (LLMs) have become the dominant workload on modern AI accelerators, yet deploying them on specialized hardware still faces two core challenges: how to import a trained model into a compiler-friendly intermediate representation, and how to efficiently schedule the autoregressive inference loop under limited on-chip memory.
This paper presents an MLIR (Multi-Level Intermediate Representation) based compilation method for large language models, illustrated using two dialects of operators, TopOp and TpuOp.
TopOp serves as a high-level graph dialect that is independent of both the source framework and the target chip, and is responsible for expressing model semantics; TpuOp serves as the target hardware dialect, carrying chip-related decisions such as quantization, layer groups, and memory layout.
A model is first represented as TopOp, then lowered layer by layer to TpuOp, and finally a deployable binary is generated.
In addition, each Transformer layer is split into three stages for static compilation: prefill, prefill\_kv (prefill with historical key-value cache), and decode, so as to accommodate the different computational characteristics of prompt-parallel processing and per-token generation.
The method has been implemented in the TPU-MLIR compiler\footnote{https://github.com/sophgo/tpu-mlir} and the LLM-TPU deployment project\footnote{https://github.com/sophgo/LLM-TPU}, supporting a variety of generative models including the Qwen, Llama, InternVL, and MiniCPM-V series, as well as multiple quantization and deployment forms such as GPTQ, AWQ, and AutoRound.

%% file: introduction.tex
Transformer-based large language models~\cite{vaswani2017attention} have evolved from research prototypes into production-grade services.
Frameworks such as PyTorch and HuggingFace Transformers~\cite{huggingface} greatly simplify training and experimentation, but the resulting model weights usually cannot run directly on specialized AI accelerators.
The traditional approach of hand-writing operator libraries for each hardware is costly and hard to keep pace with model evolution.
Therefore, the industry increasingly relies on domain-specific compilers to automate the transformation from model description to hardware instructions.

MLIR~\cite{lattner2021mlir} provides a reusable and extensible compiler infrastructure that allows developers to express computation at different levels of abstraction through custom dialects.
This paper presents an MLIR compilation method for large language models, illustrated concretely with TopOp and TpuOp: TopOp first captures framework-agnostic model semantics, which is then converted to target-specific TpuOp through a standard lowering pipeline.
To address the difference between prompt processing and token generation in autoregressive inference, each Transformer layer is statically compiled into three variants: prefill, prefill\_kv, and decode.

This paper focuses on two topics: how to import a large language model with TopOp and deploy it to an accelerator through TpuOp; and why an LLM is split into the three stages of prefill, prefill\_kv, and decode, and how each stage is represented in the compilation flow.
We intentionally keep the discussion at the methodological level; concrete implementation details (such as chip-level operator fusion, quantization tuning, and layer-group slicing) can be adjusted flexibly according to the target hardware.

%% file: background.tex
\subsection{MLIR and Multi-Level Dialects}

MLIR~\cite{lattner2021mlir} is a novel compiler infrastructure that emphasizes reusable and extensible intermediate representations.
Its core abstractions include operations, values, types, attributes, and dialects.
A dialect logically organizes a set of related operations, types, and attributes, allowing the same program to coexist at different levels of abstraction and to be progressively lowered to the target representation.

When targeting deep learning compilation, two levels of dialects are typically defined:

\begin{itemize}
\item \textbf{High-level graph dialect}: independent of the source framework and the target chip, expressing the semantics of the neural network graph, such as \op{MatMul}, \op{RMSNorm}, \op{Rope}, \op{FAttention} (short for Full Attention), \op{Reshape}, \op{Concat}, \op{MLP}, etc.
\item \textbf{Target dialect}: after determining the quantization mode, data type, memory layout, and hardware instructions, expressing the same computation. This dialect directly targets code generation.
\end{itemize}

A standard pass pipeline is responsible for lowering the high-level graph dialect operators to the target dialect operators, performing calibration when needed, running layer-group and memory planning, and finally generating the executable binary for the target hardware.

\subsection{TopOp and TpuOp Examples}

For ease of exposition, this paper uses TopOp and TpuOp as concrete examples of the two dialects of operators.

\textbf{TopOp} is the set of operators in the high-level graph dialect, used to encode the semantics of deep learning graphs.
It is independent of source frameworks such as PyTorch and TensorFlow, and also independent of any specific accelerator.
Typical TopOps include:

\begin{itemize}
\item \diop{top}{MatMul}: matrix multiplication;
\item \diop{top}{RMSNorm}: RMS normalization;
\item \diop{top}{Gather}: fetching data based on indices;
\item \diop{top}{Rope}: RoPE operation for LLMs;
\item \diop{top}{Reshape}: tensor reshape;
\item \diop{top}{Concat}: concatenating tensors along a specified axis;
\item \diop{top}{FAttention}: Full Attention (FAttention) for LLMs, encapsulating scaled dot-product attention and the causal mask;
\item \diop{top}{MLP}: the MLP operator for LLMs;
\item \diop{top}{Weight}: referencing weight data.
\end{itemize}

TopOp usually operates on tensor values of \op{RankedTensorType}, maintaining a high level of abstraction that facilitates hardware-independent graph transformations and equivalent rewrites.
Among them, \diop{top}{MLP} and \diop{top}{FAttention} are kept as fused high-level operators rather than expanded into basic operators such as \op{MatMul} + activation + residual, because their target-chip implementations usually require dedicated fused instruction sequences (such as attention fusion kernels~\cite{flashattention2022} and MLP fusion kernels).
Keeping the fusion boundary at the high level allows the lowering stage to choose the optimal fusion strategy for different chips, rather than passively matching at the granularity of basic operators.

\textbf{TpuOp} is the set of operators in the target hardware dialect, generated by the lowering pass according to the target chip characteristics after TopOp has expressed the semantics.
Compared with TopOp, TpuOp additionally carries chip-related attributes, such as:

\begin{itemize}
\item data type (F32, BF16, F16, INT8, etc.);
\item quantization format support, including symmetric/asymmetric INT8, and direct-through compilation of weight-quantized models such as AWQ / GPTQ / AutoRound;
\item layer-group information (\op{group\_info}) for on-chip memory scheduling;
\item operator splitting information (e.g., splitting by the number of cores);
\item address information (required memory size and storage offset);
\item target chip instruction encapsulation.
\end{itemize}

\begin{figure}[t]
\centering
\includegraphics[width=0.75\linewidth]{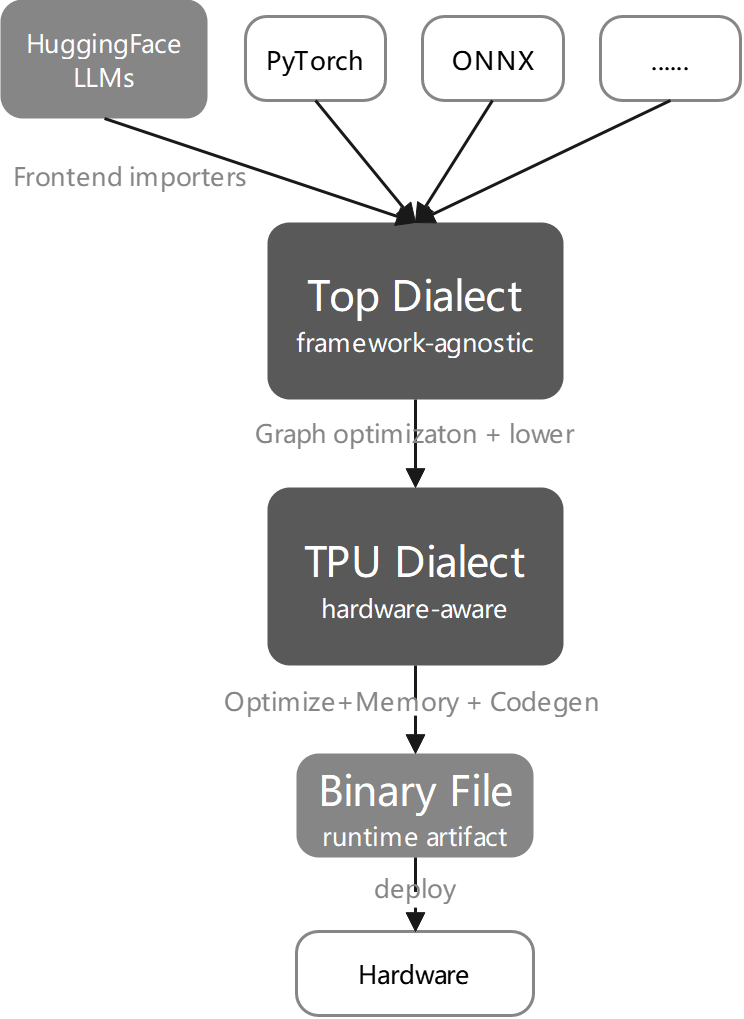}
\caption{Relationship between TopOp and TpuOp.}
\label{fig:dialects}
\end{figure}

The relationship between TopOp and TpuOp is shown in Figure~\ref{fig:dialects}.

\subsection{Overview of LLM Inference}

The rough execution process of an LLM is as follows:

\begin{enumerate}
\item Convert the input text into input tokens;
\item Obtain hidden states through word embedding;
\item Pass through \op{num\_layers} blocks to produce hidden states;
\item Obtain logits through the LM head;
\item Output a token from the logits via a sampling algorithm;
\item Repeat steps 2--5 for that token until an end-of-sequence symbol is encountered.
\end{enumerate}

\begin{figure*}[t]
\centering
\includegraphics[width=0.82\textwidth]{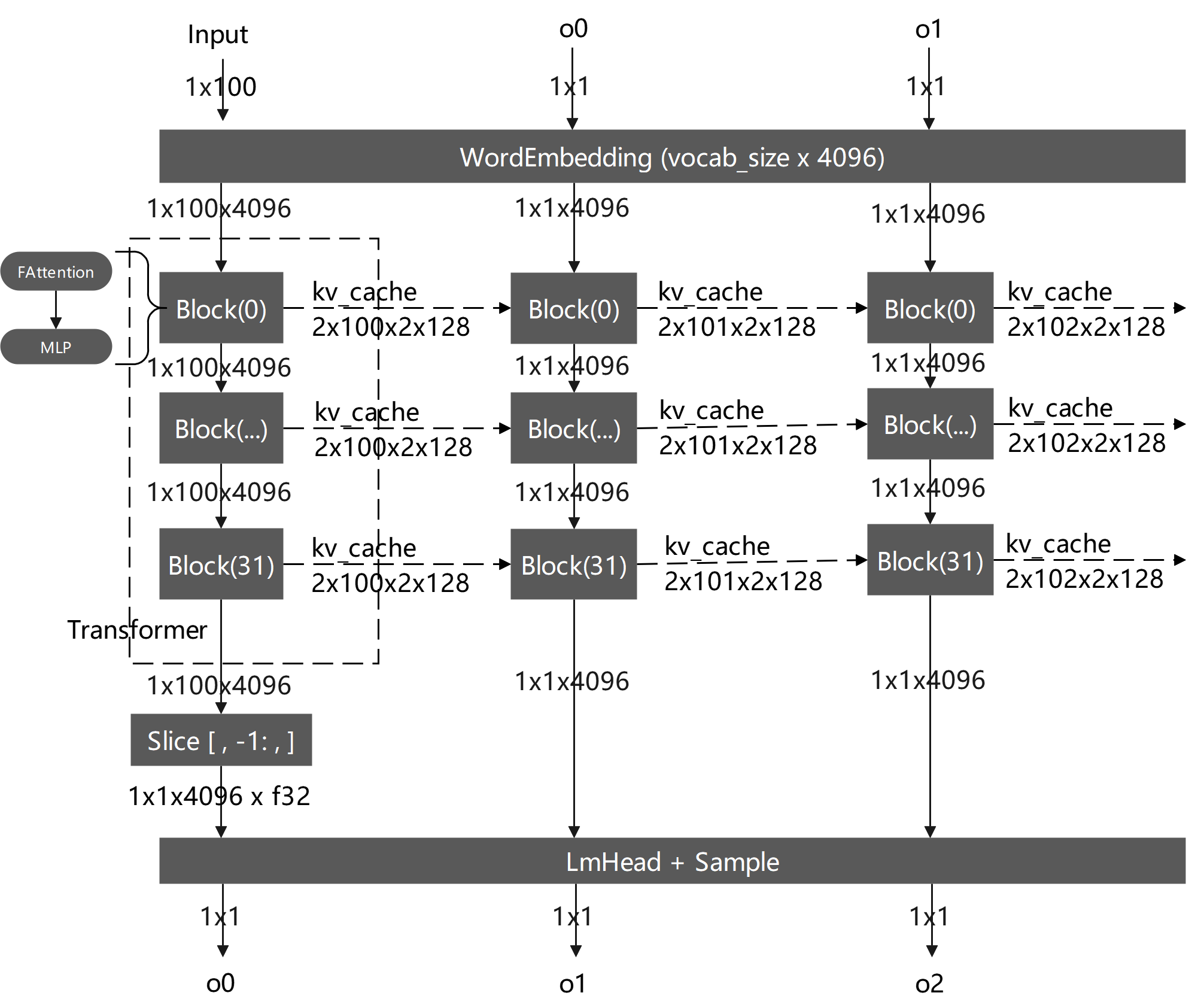}
\caption{LLM execution flow.}
\label{fig:llm_flow}
\end{figure*}

The block computation mainly consists of RoPE, Full Attention, MLP, etc., and each block generates the corresponding K cache and V cache for each token, collectively referred to as the KV Cache.
The overall execution flow of the large model is shown in Figure~\ref{fig:llm_flow}.

For ease of understanding, the figure assumes: \op{num\_layers} is 32, the number of input tokens is 100, \op{hidden\_size} is 4096, \op{kv\_head} is 2, and \op{head\_dim} is 128; o0, o1, and o2 represent the sequentially output tokens.

%% file: compiler.tex
The first topic of this paper is the complete path from a trained checkpoint to a target hardware binary.
Taking \op{llm\_convert.py} in TPU-MLIR~\cite{tpumlir2022} as an example, the script loads the HuggingFace model configuration and weight shards, dispatches to the corresponding converter according to the model family (e.g., for dense decoder-only Transformers, grouped-query attention variants, and multimodal language models, respectively), and then sequentially completes TopOp MLIR generation, lowering of each module to TpuOp, compilation, and result merging.

\subsection{Model Import and TopOp Module Generation}

Below is a TopOp pseudocode example for one block (some parameters and shape annotations are omitted for brevity).

\begin{figure*}[t]
\centering
\begin{lstlisting}[style=python-code]
# A block (without historical KV)
input   = top.InputOp("input_states", 0)
pos_ids = top.InputOp("position_ids", 1)

x   = top.RMSNorm(input)                  # input_layernorm
q   = top.MatMul(x, w_q)                    # q_proj
k   = top.MatMul(x, w_k)                    # k_proj
v   = top.MatMul(x, w_v)                    # v_proj
cos = top.Gather(rotary_cos_w, pos_ids)     # precomputed cos/sin table
sin = top.Gather(rotary_sin_w, pos_ids)
q   = top.Rope(q, cos, sin)
k   = top.Rope(k, cos, sin)
attn = top.FAttention(q, k, v)
h   = top.Add(input, top.MatMul(attn, w_o))     # o_proj + residual
o   = top.Add(h, top.MLP(top.RMSNorm(h)))     # post_attn_ln + MLP + residual
return top.Return(o, k, v)                  # output the new K/V of this layer
\end{lstlisting}
\caption*{Listing 1: A block (without historical KV).}
\end{figure*}

\begin{figure*}[t]
\centering
\begin{lstlisting}[style=python-code]
# A block (with historical KV)
input   = top.InputOp("input_states", 0)
pos_ids = top.InputOp("position_ids", 1)
k_cache = top.InputOp("history_k", 2)
v_cache = top.InputOp("history_v", 3)

x   = top.RMSNorm(input)                  # input_layernorm
q   = top.MatMul(x, w_q)                    # q_proj
k   = top.MatMul(x, w_k)                    # k_proj
v   = top.MatMul(x, w_v)                    # v_proj
cos = top.Gather(rotary_cos_w, pos_ids)     # precomputed cos/sin table
sin = top.Gather(rotary_sin_w, pos_ids)
q   = top.Rope(q, cos, sin)
k   = top.Rope(k, cos, sin)
k_all = top.Concat(k_cache, k)              # concat historical KV then attend
v_all = top.Concat(v_cache, v)
attn  = top.FAttention(q, k_all, v_all)
h   = top.Add(input, top.MatMul(attn, w_o))     # o_proj + residual
o   = top.Add(h, top.MLP(top.RMSNorm(h)))     # post_attn_ln + MLP + residual
return top.Return(o, k, v)                  # output the new K/V of this layer
\end{lstlisting}
\caption*{Listing 2: A block (with historical KV).}
\end{figure*}

Here \diop{top}{Gather} is used to fetch the precomputed cos/sin values according to \op{position\_ids}, rather than computing RoPE directly at runtime; the reason is explained in Section~\ref{sec:perf} on performance optimization.

After the frontend construction, the generated MLIR file format is roughly as follows:

\begin{figure*}[t]
\centering
\begin{lstlisting}[style=mlir-code]
func.func @block_0(%arg0: tensor<1x2048x4096xf32>, ...) -> ... {
  %0 = "top.RMSNorm"(%arg0, %weight0) ...
  %1 = "top.MatMul"(%0, %weight_q) ...
  %2 = "top.MatMul"(%0, %weight_k) ...
  %3 = "top.MatMul"(%0, %weight_v) ...
  ...
  %i = "top.FAttention"(%q, %k_cache, %v_cache, ...) ...
  ...
  %n = "top.MLP"(%m) ...
  ...
  return %hidden, %k_cache, %v_cache : ...
}
\end{lstlisting}
\caption*{Listing 3: Generated TopOp MLIR for a block.}
\end{figure*}

\subsection{Lowering from TopOp to TpuOp}

After the TopOp module is generated, the compiler converts it to TpuOp through the lowering pipeline.
During lowering, each TopOp is replaced by the corresponding TpuOp according to the target chip's quantization mode, data type, and memory constraints, and chip-related attributes are attached.
For example:

\begin{itemize}
\item After \diop{top}{MatMul} is lowered to \diop{tpu}{MatMul}, weight quantization parameters and scaling factors are attached under W4A16 / INT8 quantization modes; for already-quantized weights such as GPTQ / AWQ, the compiler can directly parse their compressed format and generate the corresponding TpuOp;
\item After \diop{top}{FAttention} is lowered to the fused attention TpuOp of the corresponding chip, the backend generates the concrete TPU instruction sequence;
\item All TpuOps may also carry \op{group\_info} (layer-group information), physical addresses, and operator splitting information, for use by subsequent layer-group scheduling, memory allocation, and code generation.
\end{itemize}

TPU-MLIR supports direct-through compilation of F32, BF16, F16, INT8 (symmetric/asymmetric), and quantized models such as AWQ / GPTQ / AutoRound.
For the INT8 mode, a calibration pass is inserted to determine the scale and zero-point of each tensor using a small amount of sample data.

After lowering is complete, the compiler continues to run standard optimization and code generation passes: layer-group slicing, on-chip/off-chip memory allocation, and codegen, finally outputting a per-module target binary.
Since a large language model cannot be processed as a single compilation unit, the frontend splits the model into multiple small TopOp modules---one module per Transformer layer variant, plus embedding, the language model head, and auxiliary heads---and compiles each module independently.
The final binary is obtained by merging all per-module results.
This modular approach has two advantages: compilation can be parallelized across layers; and at runtime the functions can be invoked in the order required by autoregressive generation.

\begin{figure}[t]
\centering
\includegraphics[width=0.9\linewidth]{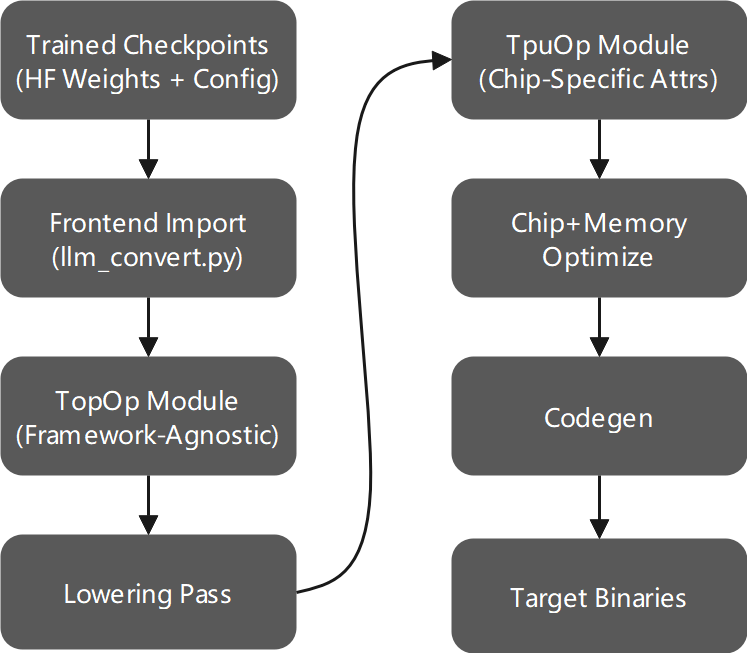}
\caption{Overall compilation pipeline (checkpoint $\to$ TopOp $\to$ TpuOp $\to$ binary).}
\label{fig:compile_flow}
\end{figure}

The overall compilation pipeline from a trained checkpoint to a deployment binary is shown in Figure~\ref{fig:compile_flow}.

\begin{figure*}[t]
\centering
\includegraphics[width=0.75\textwidth]{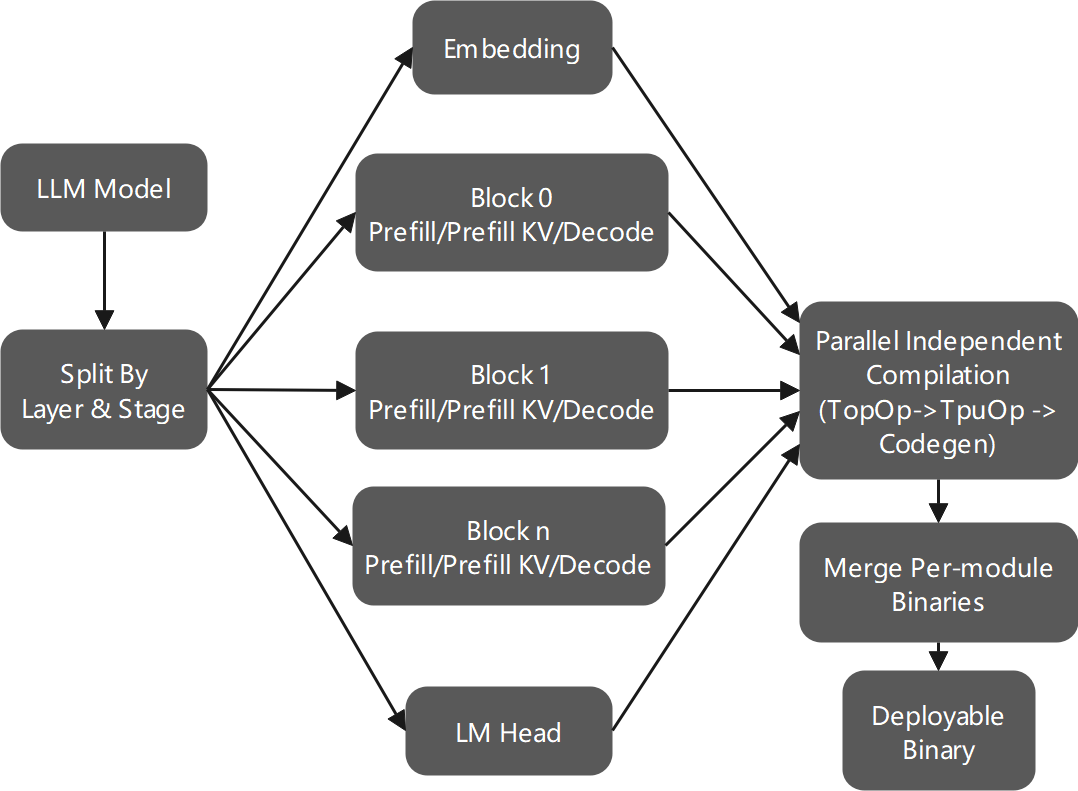}
\caption{Modular splitting, parallel compilation, and merging.}
\label{fig:parallel}
\end{figure*}

The process of modular splitting, parallel compilation, and merging is shown in Figure~\ref{fig:parallel}.
The frontend splits the model into multiple small TopOp modules by layer and by stage; each module independently completes TopOp $\to$ TpuOp $\to$ codegen, and the results are merged into the final deployment binary.

%% file: stages.tex
The second topic of this paper is the execution model.
Autoregressive LLM inference is not a single forward pass, but alternates between processing a (possibly long) prompt and generating tokens one by one.
Therefore, we split each Transformer layer into three TopOp module variants for static compilation: prefill, prefill\_kv, and decode, and then lower them to the corresponding TpuOp variants respectively.

\subsection{Why Split into Three Stages?}

\begin{itemize}
\item \textbf{Prefill}: processes all tokens in the input prompt in parallel. The query length equals the prompt length, and a Key/Value tensor is generated for each token.
\item \textbf{Decode}: processes only the latest generated token, but must attend to all previously computed Keys and Values. The query length is 1, while the Key/Value length grows with the generation sequence.
\item \textbf{Prefill\_kv}: used when the runtime wants to continue a conversation and needs to merge the historical Key/Value cache with a new prompt. Its behavior is similar to prefill, but it first concatenates the historical cache before the new tokens.
\end{itemize}

The main reason for splitting into three stages is that the tensor shapes and memory layouts of these three cases differ significantly; if compiled into a single function, a large amount of padding or dynamic recompilation would be forced.
Most specialized AI accelerators are parallel architectures oriented toward tensor operations and rely heavily on static compilation; dynamic compilation not only increases runtime compilation overhead, but also reduces the determinism of memory allocation and operator scheduling.
To ensure that as many parts as possible can be statically compiled, prefill and prefill\_kv are usually precompiled at a fixed maximum length (e.g., 8K).
By generating independent TopOp modules, each stage can be statically optimized for its own tensor shapes, then lowered to a dedicated TpuOp implementation, with memory allocated statically and codegen more deterministic, thereby reducing dynamic overhead at runtime.

Splitting into three stages can simultaneously support three typical scenarios:

\begin{enumerate}
\item When historical context is not considered, use \textbf{Prefill} + \textbf{Decode};
\item When historical context is supported, the first session uses \textbf{Prefill} + \textbf{Decode}, and subsequent sessions use \textbf{Prefill\_kv} + \textbf{Decode};
\item When the prompt length is very large, e.g., prefill and prefill\_kv are both precompiled at 8K while the prompt reaches 80K, it can be completed in segments using 1 Prefill + 9 Prefill\_kv.
\end{enumerate}

The main differences among the three stages are summarized in Table~\ref{tab:stages}.

\begin{table*}[t]
\centering
\caption{Comparison of the prefill, prefill\_kv, and decode stages. Let $h$~=~hidden\_states, $p$~=~position\_ids, $m$~=~attn\_mask, $hk/v$~=~history\_k/v; all three stages output $(h, k/v\ \text{cache})$.}
\label{tab:stages}
\small
\renewcommand{\arraystretch}{1.15}
\begin{tabular}{llcc>{\raggedright\arraybackslash}p{0.30\linewidth}}
\toprule
Stage & Input tokens & Hist. KV & Input & \multicolumn{1}{c}{Typical use} \\
\midrule
Prefill     & $>1$ & No  & $h, p$          & First-round prompt encoding \\
Prefill\_kv & $>1$ & Yes & $h, p, hk/v$    & Multi-turn continuation / long-prompt segmentation \\
Decode      & $1$  & Yes & $h, p, m, hk/v$ & Per-token generation \\
\bottomrule
\end{tabular}
\end{table*}

\subsection{Prefill Stage}

Prefill and prefill\_kv are actually generated from the same parameterized TopOp template, distinguished only by a \op{with\_history} flag: when \op{with\_history=False}, the historical KV inputs are empty tensors, yielding the prefill module \op{block\_\{i\}}; when \op{with\_history=True}, the historical KV inputs carry the real cache, yielding the prefill\_kv module \op{block\_kv\_\{i\}}.
The two share the same dataflow and lowering path, avoiding duplicate implementation.
This subsection first describes prefill, whose signature is:

\begin{lstlisting}[style=sig-code]
Input:  [hidden_states, position_ids]
Output: [hidden_states, key_cache, value_cache]
\end{lstlisting}

where \op{hidden\_states} has shape \op{[1, max\_input\_length, hidden\_size]}, and \op{history\_k} / \op{history\_v} are empty tensors in pure prefill and carry the historical cache in prefill\_kv.
The Key/Value cache returned by this stage is then passed to the decode stage.

\begin{figure}[t]
\centering
\includegraphics[width=0.9\linewidth]{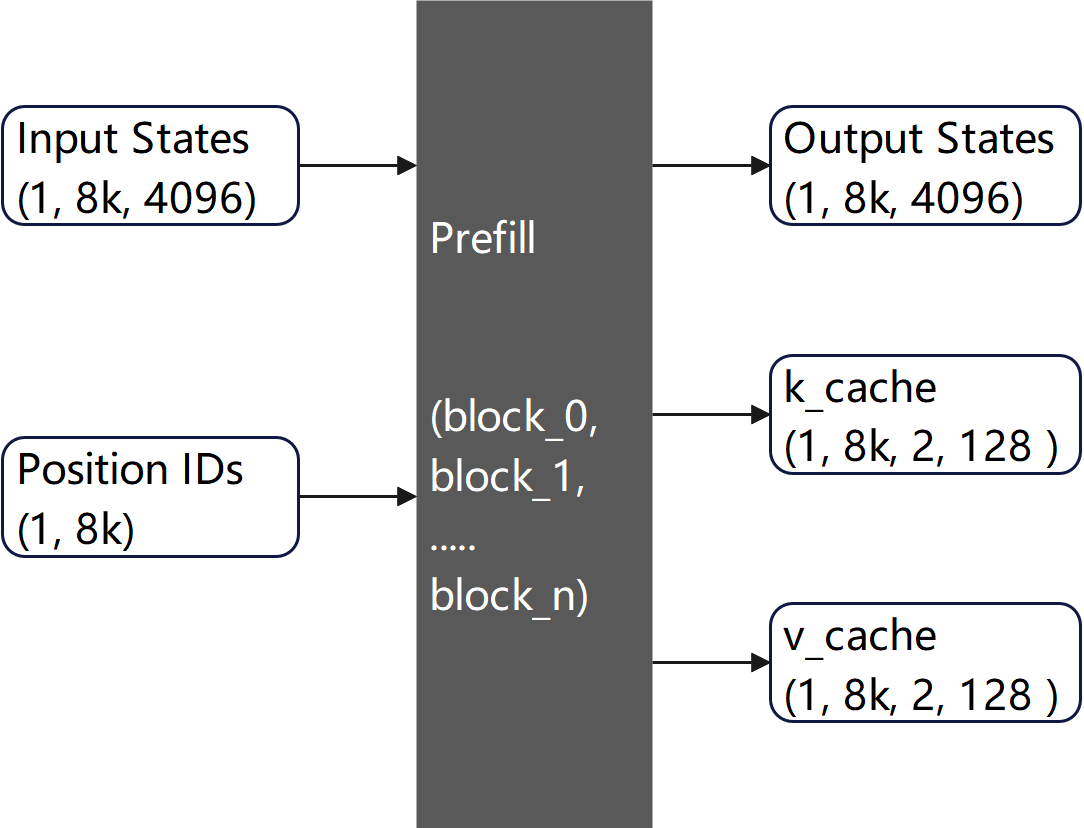}
\caption{The prefill stage (\op{block\_\{i\}}), illustrated with \op{hidden\_size}~=~4096 and a precompiled prompt length of 8K.}
\label{fig:prefill}
\end{figure}

Figure~\ref{fig:prefill} illustrates the prefill stage with a concrete example of \op{hidden\_size}~=~4096 and a precompiled prompt length of 8K.

\subsection{Prefill\_kv Stage}

\begin{figure}[t]
\centering
\includegraphics[width=0.9\linewidth]{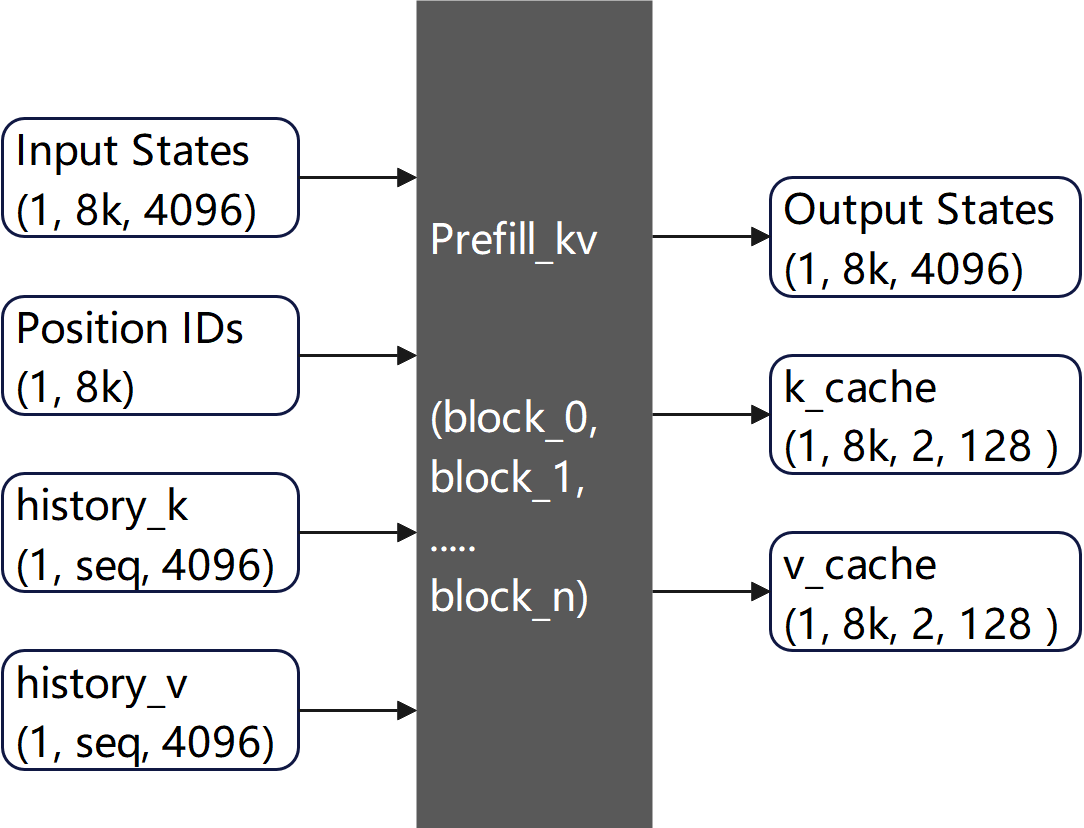}
\caption{The prefill\_kv stage (\op{block\_kv\_\{i\}}); \op{seq} denotes the maximum context length, which is fixed at compile time.}
\label{fig:prefill_kv}
\end{figure}

The prefill\_kv module \op{block\_kv\_\{i\}} is the instance of the above template with \op{with\_history=True}; its TopOp dataflow is identical to prefill, and the only difference is that the input \op{history\_k} / \op{history\_v} carry the historical cache:

\begin{lstlisting}[style=sig-code]
Input:  [hidden_states, position_ids,
         history_k, history_v]
Output: [hidden_states, key_cache, value_cache]
\end{lstlisting}

Inside the module, the newly computed Key/Value tensors are concatenated with the historical cache along the sequence dimension via \diop{top}{Concat}.
The effective Key/Value length is therefore \op{seq\_length + max\_input\_length}.
This stage is generated only when historical KV reuse is enabled, which is a common setting for long-context conversation applications; its dataflow is illustrated in Figure~\ref{fig:prefill_kv}.

\subsection{Decode Stage}

\begin{figure}[t]
\centering
\includegraphics[width=0.9\linewidth]{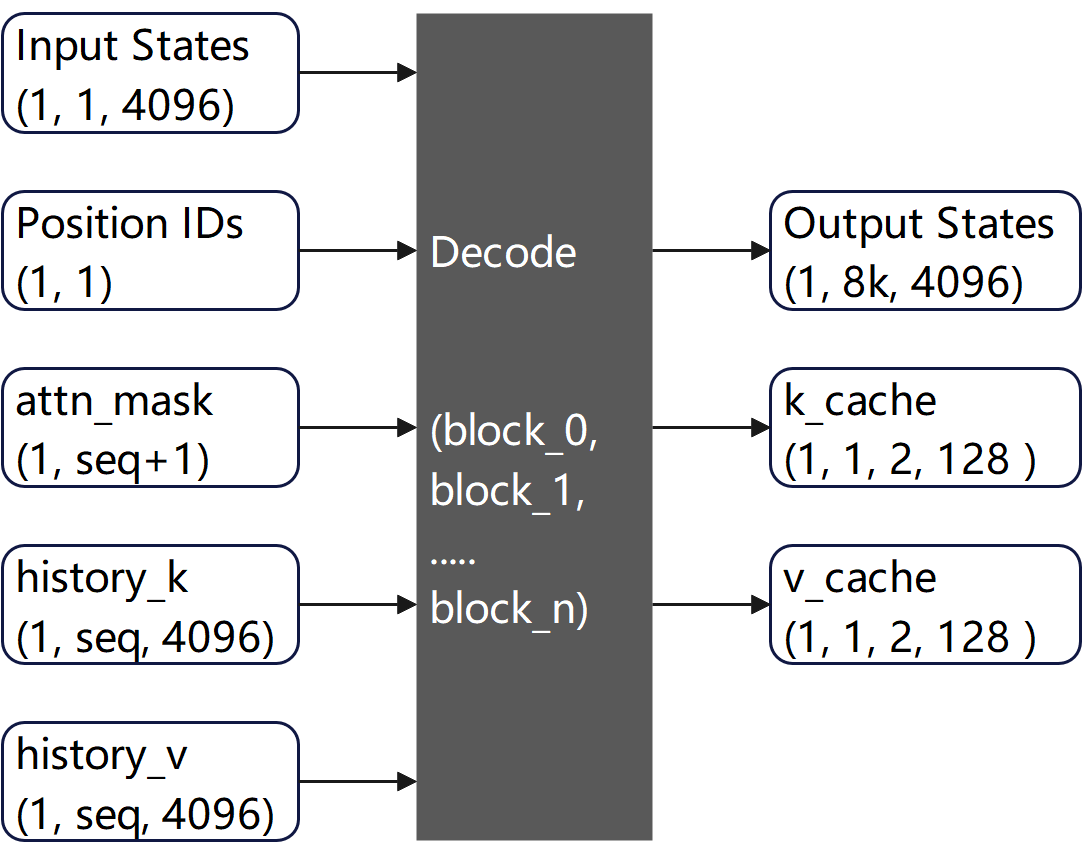}
\caption{The decode stage (\op{block\_cache\_\{i\}}), which processes one token per call and takes the attention mask as an external input.}
\label{fig:decode}
\end{figure}

The decode module \op{block\_cache\_\{i\}} processes one new token per forward call:

\begin{lstlisting}[style=sig-code]
Input:  [hidden_states, position_ids, attn_mask,
         history_k, history_v]
Output: [hidden_states, key_cache, value_cache]
\end{lstlisting}

Here \op{hidden\_states} has shape \op{[batch, 1, hidden\_size]}, and the Key/Value of the new token are concatenated into the history.
Unlike the two prefill variants, the decode stage takes an explicit \op{attn\_mask} input, as shown in Figure~\ref{fig:decode}: the mask is filled in by the runtime according to the current context length and passed in from the outside, so that the module itself keeps a fully static shape --- again in the interest of static compilation.
Since the context length keeps growing during generation, the compiler may optionally generate multiple decode variants for different cache lengths (i.e., a ``decode chunk list''), and at runtime select the variant whose static shape best matches the current context size.

\subsection{Runtime Orchestration}

\begin{figure}[t]
\centering
\includegraphics[width=0.55\linewidth]{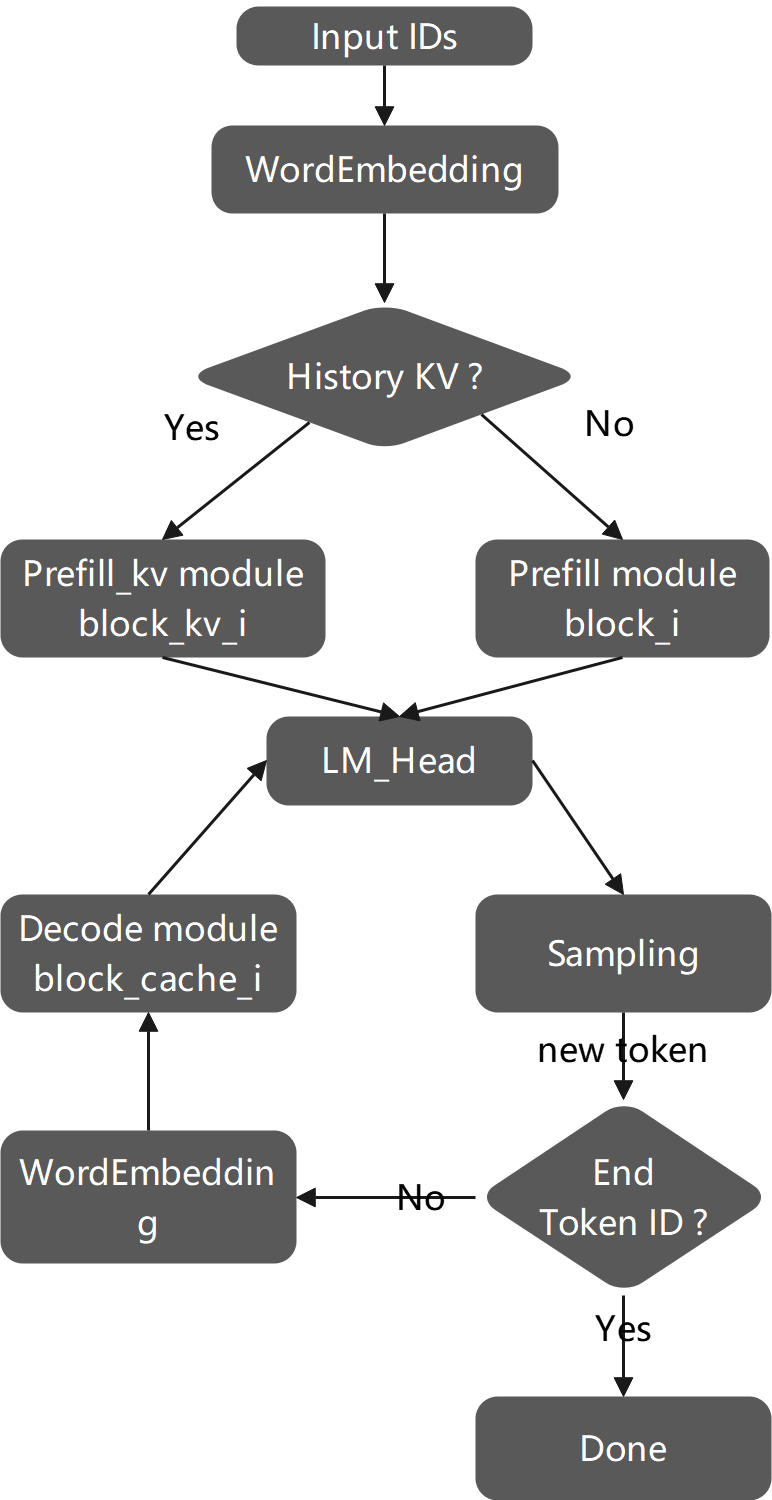}
\caption{Three-stage runtime orchestration of autoregressive inference.}
\label{fig:runtime}
\end{figure}

The three stages are not mutually exclusive, but are used in sequence.
A typical inference loop is shown in Figure~\ref{fig:runtime}.

A textual description of the typical inference loop is as follows:

\begin{enumerate}
\item Run the embedding module on the prompt token ids.
\item For each layer, run the \textbf{prefill} (or \textbf{prefill\_kv} if history exists) module to obtain the Key/Value cache and the hidden state at the end of the prompt.
\item Run the language model head to obtain the logits of the last prompt token.
\item Sample to obtain the next token.
\item Run the \textbf{decode} module with the ever-growing Key/Value cache to obtain the hidden state of the new token.
\item Repeat steps 3--5 until an end-of-sequence symbol is generated.
\end{enumerate}

Since each stage is an independently compiled function in the merged deployment binary, the runtime can freely combine them.
The static shapes inside the functions keep the generated hardware instructions simple, while the coarse-grained split between stages captures the dynamic nature of autoregressive generation.

\subsection{Typical Performance Optimizations}\label{sec:perf}

Taking a specialized AI accelerator as an example, we present several common optimization ideas; the specific parameters need to be adjusted according to the target chip and model size.

\subsubsection{Causal Mask Support}

By algorithmic logic, the attention of an 8K-length input requires an $8K \times 8K$ causal mask, and loading it directly wastes a lot of bandwidth.
A fixed small mask (e.g., $128 \times 128$) can be used; as long as the tile size of the full attention (e.g., $64 \times 64$, determined by the number of accelerator cores and the on-chip cache size) is fixed, this small mask can be reused, significantly reducing mask bandwidth.
This is also one of the reasons why the prefill / prefill\_kv modules do not need to take \op{attention\_mask} as an explicit input.
The decode module, by contrast, does take a small \op{attn\_mask} as an external input; since the runtime rather than the module fills in the mask, the decode module itself remains free of mask-generation logic and keeps a static shape.

\subsubsection{Implementation of Positional Encoding}

By algorithmic logic, positional encoding involves the cos/sin series operations required by RoPE, and such element-wise operations are inefficient on specialized accelerators.
We can precompute the position vectors from $0$ to \op{seq\_length} at compile time and store them as weights, and at runtime fetch the corresponding values via \diop{top}{Gather} according to \op{position\_ids}, thereby turning runtime computation into memory access.
This scheme trades a constant storage of \op{seq\_length $\times$ head\_dim $\times$ 2} floating-point numbers for the elimination of element-wise computation, and is usually a net gain in the decode stage where on-chip compute is more likely to be the bottleneck than memory access.

\subsubsection{KV Cache Management}

The compiler needs to specially plan the memory addresses of the KV Cache.
Whereas mainstream serving systems such as vLLM~\cite{vllm2023} manage the KV cache dynamically with paged attention, our static-compilation setting requires the cache layout to be fixed at compile time.
In particular, when using \diop{top}{Concat} to concatenate the KV of a new token into the historical cache, it should be implemented through address reuse or in-place updates to avoid extra bandwidth consumption from physical copies.
In the decode implementation of LLM-TPU, the corresponding \op{block\_cache\_*} variant is also selected according to the current context length, so that the address allocation and access pattern of the KV Cache remain statically predictable.

With the above optimizations, a specialized AI accelerator can fully exploit compute in the prefill stage and fully utilize off-chip bandwidth in the decode stage, thereby achieving high hardware utilization for overall inference.

%% file: experiments.tex
We evaluate the proposed compilation method on the Qwen3.5 model
family, deployed through the TPU-MLIR / LLM-TPU toolchain described in
this paper.

\subsection{Setup}
All experiments run on a single Sophgo BM1684X edge SoC, which provides
16\,TFLOPS of FP16 peak compute and 64\,GB/s of DDR memory bandwidth.
Models are converted with \texttt{llm\_convert.py}, which lowers each
Transformer layer into the \textit{prefill}, \textit{prefill\_kv}, and
\textit{decode} stages of Sec.~\ref{sec:stages} and emits a deployable
binary. All models use W4BF16 quantization (4-bit weights with BF16
activations) and are validated at a 2K context length. We report
end-to-end decode throughput in tokens per second (TPS), measured over
the full inference pipeline --- including token encoding and decoding on
the host CPU --- together with the resulting memory-bandwidth
utilization (BW Util), i.e., the ratio of the effective bandwidth
(TPS $\times$ model size) to the peak DDR bandwidth.

The Qwen3.5 models are vision-language variants with a ViT front end;
for them the ViT and the prefill stage are compiled with dynamic shapes,
while decode remains fully static.

\subsection{Results}
Table~\ref{tab:perf} summarizes the measurements.

\begin{table}[t]
  \centering
  \small
  \caption{Decode throughput (tokens/s) and memory-bandwidth utilization
  of Qwen3.5 models on the BM1684X edge SoC (16\,TFLOPS FP16, 64\,GB/s
  DDR) under W4BF16 quantization at a 2K context length. Throughput is
  end-to-end, including host-side token encoding and decoding.}
  \label{tab:perf}
  \begin{tabular}{lrrr}
    \toprule
    Model & Size (GB) & TPS & BW Util \\
    \midrule
    Qwen3.5-2B & 1.67 & 30.00 & 78.0\% \\
    Qwen3.5-4B & 3.00 & 16.77 & 78.6\% \\
    Qwen3.5-9B & 5.39 & 9.84  & 83.0\% \\
    \bottomrule
  \end{tabular}
\end{table}

\textbf{Decode throughput is bandwidth-bound and scales with weight
size.}
Because the decode stage reads all weights once per generated token, TPS
is approximately inversely proportional to the on-disk model size: as
the model grows from 1.67\,GB to 5.39\,GB ($3.2\times$), throughput
drops from 30.0 to 9.84 tokens/s ($3.0\times$). This confirms that the
quantization decisions carried by the TpuOp dialect directly translate
into decode-side gains, exactly as the bandwidth model predicts;
aggressive W4BF16 quantization is also what makes a 9B-class model fit
the device at all.

\textbf{The achieved bandwidth utilization is consistently high.}
Across all three model sizes, the decode stage sustains 78--83\% of the
peak DDR bandwidth --- a high level for an edge device. These numbers
cover the full pipeline, including token encoding and decoding performed
on the host CPU; the utilization of the chip-side inference alone is
even higher. The slight upward trend with model size reflects the
shrinking relative cost of the fixed per-token overheads.

\textbf{The results validate the flow on multimodal models.}
Qwen3.5 models couple a ViT image encoder with the LLM; our toolchain
compiles the ViT and the prefill stage with dynamic shapes while keeping
decode fully static, and still reaches 78--83\% bandwidth utilization.
This shows that the three-stage compilation scheme composes cleanly with
additional front-end networks.

Overall, the results show that the proposed TopOp$\to$TpuOp lowering
flow, combined with per-stage static compilation, delivers practical
interactive throughput (30 tokens/s at the 2B scale and $\sim$10
tokens/s at the 9B scale) at 78--83\% of peak memory bandwidth on a
64\,GB/s edge SoC.

%% file: discussion.tex
The two topics of this paper are tightly coupled.
TopOp provides a common expression language for the model, while the three-stage splitting is the strategy that cuts this common representation into hardware-specific TpuOp modules that can be executed efficiently at runtime.

The following design decisions are worth emphasizing:

\begin{itemize}
\item \textbf{Framework independence.} TopOp modules contain no PyTorch or HuggingFace concepts, only tensor operations. This allows the same TopOp $\to$ TpuOp lowering pipeline to apply to dense decoder-only Transformers, grouped-query attention variants, mixture-of-experts variants, and vision-language models.
\item \textbf{Modular compilation.} Splitting the model into TopOp modules by layer and by stage keeps each compilation unit small, supports parallel compilation, and allows the runtime to reuse compiled layers under multiple sequence lengths.
\item \textbf{Per-stage static shapes.} Rather than compiling a fully dynamic Transformer, we compile several static-shape variants and schedule among them at runtime. This aligns with MLIR's design philosophy of encoding transformation validity preconditions directly into the IR.
\end{itemize}

Future work includes extending the three-stage model to speculative decoding, supporting longer contexts through more aggressive cache tiling, and applying the same splitting method to multimodal language models whose vision tower output feeds into the same prefill/decode loop.

%% file: conclusion.tex
This paper presented an MLIR-based compilation method for large language models, illustrated with TopOp and TpuOp.
A model is first imported into the framework-agnostic TopOp, then lowered to the target-hardware-specific TpuOp, and compiled into deployment functions partitioned by stage.
To handle autoregressive inference, each Transformer layer is split into three static-shape stages: prefill, prefill\_kv, and decode.
This design maintains the compiler's generality across Transformer families while providing the runtime with dedicated functions for efficiently handling prompts and per-token generation.

The above method has been validated in a real compiler project.
Based on this method, we implemented LLM support in the reference project TPU-MLIR\footnote{https://github.com/sophgo/tpu-mlir}, and provided LLM-TPU\footnote{https://github.com/sophgo/LLM-TPU}, a demo project supporting a variety of large models, to demonstrate the complete flow from model import to deployment.
On the Sophgo BM1684X edge SoC, Qwen3.5 models compiled with this method sustain 78--83\% of the peak DDR bandwidth during decode, confirming that the statically compiled three-stage design achieves high hardware utilization in practice.